\documentclass{article}
\usepackage{iclr2027_conference, times}
\usepackage{hyperref}
\usepackage{url}
\usepackage{amsthm,amsmath,amssymb}
\usepackage{mathrsfs}
\usepackage{dsfont}
\usepackage{amsthm}
\usepackage{multirow}
\usepackage{multicol}
\usepackage{float}
\usepackage{booktabs}
\usepackage{xcolor}
\usepackage[table]{xcolor}
\usepackage{arydshln}
\theoremstyle{plain}

\usepackage{graphicx}
\usepackage{subcaption}

\usepackage{pifont}

\usepackage{wrapfig}
\usepackage{caption} 

\theoremstyle{remark}

\title{SpatialSkill: Self-Evolving Skills for Cross-View Spatial Reasoning}

\author{%
  Ruifan Zuo$^{1}$\quad
  Guocheng Hu$^{1}$\quad
  Wanshui Gan$^{2}$\quad
  Junyi Wang$^{1}$\quad
  Xiang Lei$^{3}$\quad
  Tian Gan$^{1\dagger}$ \\
  $^{1}$Shandong University \quad
  $^{2}$Shanghai AI Laboratory \quad
  $^{3}$Zhiyang Innovation Co., Ltd. \\
  $^{\dagger}$ Corresponding Author\\
  {\tt gantian@sdu.edu.cn}
}

\iclrfinalcopy

\begin{document}

\maketitle

\begin{abstract}
Cross-view spatial reasoning requires a model to align different viewpoints into a coherent spatial representation, yet this ability remains challenging for vision-language models despite being natural to humans.
Existing methods typically improve spatial reasoning by updating model weights, which keeps the acquired knowledge implicit and tied to a specific backbone. 
We propose \textit{SpatialSkill}, a weight-update-free framework that enables a frozen vision-language model to accumulate explicit natural-language reasoning skills from offline trajectories. 
Unlike symbolic tasks, perceptual skills cannot be reliably verified simply by executing them: a plausible spatial rule may lack visual support or require transformations that the frozen model cannot perform.
SpatialSkill therefore admits candidate skills only after visual-grounding and executability checks, constrains manual evolution to prevent harmful regressions, and routes skills by spatial-reasoning category to reduce negative transfer. 
On CityCube, across four frozen executors, SpatialSkill yields consistent gains, and a 9B executor equipped with SpatialSkill surpasses the strongest closed-source reference in our evaluation. 
The skills are stored in a versioned natural-language manual, making the reasoning strategies explicit and auditable without modifying model parameters.
Code at https://github.com/vindahi/SpatialSkill.
\end{abstract}
\section{Introduction}
\begin{figure}[ht]
    \centering 
    \includegraphics[width=0.95\textwidth]{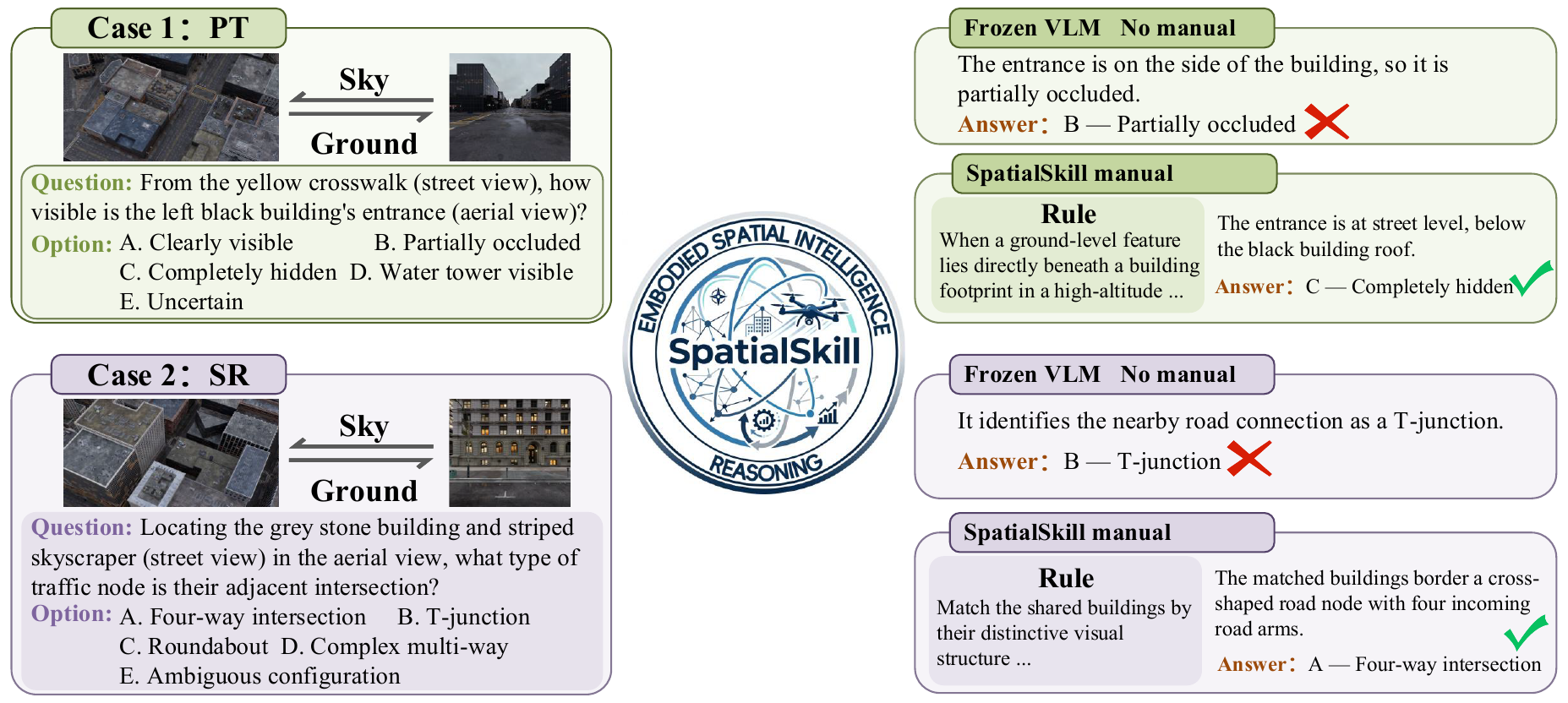} 
    \caption{A perspective-taking (PT) instance and a spatial-relation (SR) instance, each pairing aerial and ground views. Frozen executor fails on both without manuals, while a visually grounded rule from the evolved SpatialSkill manual enables it to correct its prediction without any weight update.}
    \label{fig:intro} 
\end{figure}

Spatial intelligence \citep{spatial1, omnispatial, spatial3} is central to embodied cognition. 
Missions such as UAV inspection and disaster response require relating objects, directions, and structures across aerial and ground-level observations.
We refer to this ability as cross-view spatial reasoning \citep{viewrea1, viewrea2, viewrea3, viewrea4, viewrea5}. 
Solving these cross-view problems often requires mentally transforming and aligning viewpoints rather than relying only on visual similarity or linguistic priors, yet current vision-language models remain weak at these operations \citep{spatialdise, proors2, proors3, proors4, proors5}. 
Existing approaches \citep{trainings1, trainings2, trainings3} typically improve spatial reasoning through parameter or representation updates, which encode the acquired knowledge implicitly in a particular model and make the resulting reasoning strategies difficult to inspect or revise independently of the model. 
We therefore ask: \textbf{can a frozen VLM improve its cross-view spatial reasoning without parameter updates, simply by accumulating explicit natural-language skills?} 
Fig.~\ref{fig:intro} illustrates that even a simple visually grounded rule can correct an error made by a frozen executor. 
The key challenge is therefore not whether explicit skills can help, but how to acquire, maintain, and deploy them reliably.

This raises three key challenges.
\textbf{\ding{202} Reliable skill admission.} 
Unlike symbolic skills that can be validated through direct execution, a natural-language spatial rule lacks a direct rule-level verifier and may sound plausible even when unsupported by visual evidence \citep{challenge11, challenge12, challenge13}. 
Moreover, a geometrically valid rule may still require reasoning operations beyond what the frozen executor can reliably perform. 
Thus, candidate skills must be checked for both visual grounding and executor executability before entering the skill library.
\textbf{\ding{203} Stable skill evolution.} 
Even when individual candidate skills are plausible, repeated revisions may accumulate noise, interact unexpectedly, or remove useful rules during compression, gradually degrading the skill library over time \citep{challenge21, challenge22}.
Reliable skill evolution therefore requires controlling growth and preventing harmful updates from being propagated.
\textbf{\ding{204} Category-dependent skill effectiveness.} 
Different spatial-reasoning categories involve different reasoning patterns and operations, so the same manual or inference strategy may not benefit every category equally \citep{challenge31, challenge32, challenge33, challenge34}. 
Uniformly injecting skills can therefore introduce negative transfer, motivating category-conditioned deployment.

Motivated by these challenges, we introduce \textbf{\emph{Self-Evolving Skills for Cross-View Spatial Reasoning (SpatialSkill)}}, a framework that improves a frozen VLM by accumulating explicit natural-language skills without updating its parameters. 
SpatialSkill consists of three core components. 
First, a grounded admission gate admits candidate skills only after checks for visual support and compatibility with the frozen executor's capabilities. 
Second, stability controls regulate skill accumulation by limiting uncontrolled growth, protecting stable skills from blind removal, and rejecting updates that cause excessive performance degradation. 
Third, category-conditioned routing selects how the final manual is used for each spatial reasoning category, reducing negative transfer from uniform skill deployment. 
Together, these components produce a versioned and auditable skill manual that can be selectively deployed while keeping the executor unchanged.

Our contributions are threefold. 
(i) To our knowledge, we are the first to study self-evolving explicit natural-language skills for frozen VLMs in real-world cross-view spatial reasoning, identifying three requirements for reliability: grounded admission, stable accumulation, and category-conditioned deployment.
(ii) We propose SpatialSkill, a framework that addresses these requirements through grounded admission, stability-constrained evolution, and category-conditioned routing, storing acquired reasoning strategies in an explicit, versioned, and auditable natural-language manual without parameter updates.
(iii) On CityCube across four frozen executors, SpatialSkill consistently yields accuracy gains of 2.65--4.92 percentage points. With SpatialSkill, the 9B executor reaches 57.58\%, outperforming the strongest closed-source baseline evaluated in our experiments.

\section{Related Work}

\paragraph{Spatial Reasoning in Vision-Language Models.}
Spatial reasoning is a central diagnostic axis for vision-language models. 
Recent benchmarks examine complementary capabilities, including egocentric video~\citep{vsibench}, multi-viewpoint localization~\citep{viewspatial}, multi-image intelligence~\citep{mmsibench}, cognitive-psychology axes~\citep{omnispatial}, unified cross-task evaluation~\citep{spatialdise}, and fine-grained failure analysis~\citep{mindthegap2025, spatialsurvey}. 
Across these settings, models remain challenged when reasoning requires viewpoint transformation and cross-view alignment rather than language priors or appearance matching. 
Recent methods address these limitations through several forms of model adaptation. 
Reinforcement learning improves grounded multi-step 3D reasoning~\citep{spatialthinker}, activates latent abilities~\citep{actial2025}, and extends to multi-view transformations~\citep{starr12026}. 
Supervised approaches inject depth signals~\citep{spatialbot} or align implicit representations~\citep{spatialforcing}, while robot-scale pretraining embeds spatial structure~\citep{spatialvla}. 
Together, these studies establish the importance of explicit spatial structure and learned adaptation for cross-view reasoning, with most improvements encoded in model weights or representations.

\paragraph{Self-Evolving Agents and Skill Accumulation.}
Self-improvement from practice provides a complementary direction for turning experience into reusable knowledge~\citep{seasurvey}. 
Existing approaches develop skill libraries that compile procedures into APIs~\citep{zhao2025spacemind, skillweaver, walt, tian2026self}, recursive augmentation with RL~\citep{skillrl, liu2026self}, resource-to-library conversion~\citep{skillfoundry, emboskill}, structured experience notes~\citep{amem}, and OS-style hierarchies~\citep{memoryos}. 
Many of these systems evolve executable artifacts or model parameters, allowing environments or training loops to provide direct evaluation signals. 
A close text-playbook precedent is ACE~\citep{ace2025}, which accumulates reusable lessons under text-task scoring. 
These two research directions leave a less explored intersection: evolving explicit natural-language skills for frozen VLMs in perceptual cross-view reasoning. 
A skill may lack visual support, exceed the executor's capabilities, or interact negatively with previously accumulated guidance. 
This setting therefore raises questions of grounded admission, stable accumulation, and selective deployment, which motivate SpatialSkill.
\section{Methodology}
\subsection{Problem Definition and Framework Overview}
We study cross-view spatial reasoning in a multiple-choice setting. 
Each example is defined as $x=(I_{1:k}, q, \mathcal{O})$, where $I_{1:k}$ represents $k$ distinct views of a scene, $q$ is a question, and $\mathcal{O}$ is the candidate answer set. 
A frozen executor predicts an answer $\hat{y} \in \mathcal{O}$. 
Each example belongs to one of five spatial reasoning categories: mental reconstruction (MR), perspective taking (PT), spatial relation reasoning (SR), world knowledge (WK), and comprehensive reasoning (CR), which are used for both category-wise evaluation and routing. 
During skill evolution, training examples additionally provide reference reasoning trajectories used as attribution references during reflection. 
SpatialSkill distills reusable reasoning patterns from these trajectories into natural-language skills, which are organized into a structured manual to guide the unchanged executor at inference.

\begin{figure}[ht]
  \centering
  \includegraphics[width=0.98\textwidth]{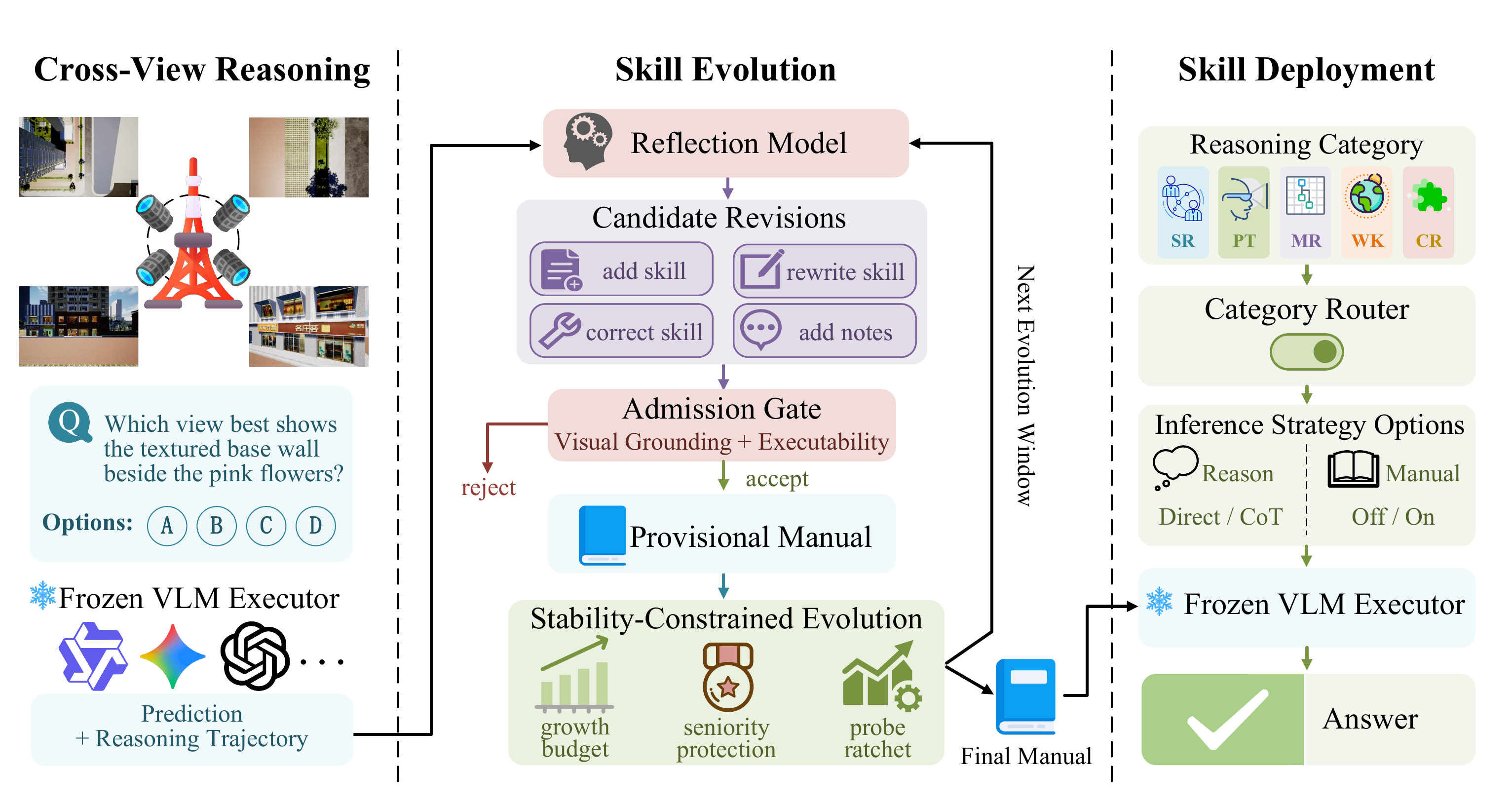}
  \caption{Overview of Self-Evolving Skills for Cross-View Spatial Reasoning (SpatialSkill)}
  \label{fig:main}
\end{figure}

SpatialSkill converts offline reasoning trajectories from a frozen VLM into a reusable natural-language skill manual and selectively deploys that manual at inference.
Fig. \ref{fig:main} summarizes this process in three stages: the frozen VLM executor first produces cross-view reasoning trajectories; a skill-evolution module reflects on these trajectories and updates the manual under grounding and stability constraints; finally, a category router determines how the manual is used for each reasoning category.

\subsection{Skill Reflection and Grounded Admission}
Reasoning trajectories capture useful experience from the frozen executor, but remain instance-specific. 
SpatialSkill therefore uses reflection to abstract reusable revisions to a natural-language skill manual. 
Because reflected candidates may still lack visual support or require operations beyond the executor's capabilities, they are screened for visual grounding and executability before entering the manual.
We maintain a skill manual $M$, initialized empty and organized into four sections: planning, cross-view alignment, spatial reasoning, and failure avoidance, together with execution notes. 
To evolve the manual incrementally, we process the evolution stream in windows. 
For window $w$, let $T_w$ denote the collected trajectories and $M_{w-1}$ the current manual. 
A fixed reflection model $\Phi$ proposes candidate revisions:
\begin{equation}
\Delta_w \;=\; \Phi\big(T_w,\, M_{w-1}\big),
\label{eq:revision}
\end{equation}
where $\Delta_w$ denotes the proposed revisions. 
Each revision performs one of four operations: adding a skill, rewriting a skill, correcting a skill, or adding an execution note. 
Each candidate is further associated with skill-level or execution-level attribution estimated against the reference trajectories. 
Execution-level revisions only update execution notes, preventing episodic noise from contaminating reusable skill rules.

Because reflected candidates are not necessarily reliable, each candidate rule $r$ is screened by an admission gate, implemented by the fixed reflection model, that checks visual grounding and executability:
\begin{equation}
\mathcal{G}(r) \;=\; \mathcal{V}(r) \ \wedge\ \mathcal{E}(r),
\label{eq:gate}
\end{equation}
where $\mathcal{V}(r)=1$ requires the rule to be supported by visual evidence, while $\mathcal{E}(r)=1$ requires the prescribed operation to remain within the frozen executor's capabilities. 
Only candidates satisfying $\mathcal{G}(r)=1$ are admitted into the manual. 
Rejected candidates are logged with their triggering evidence, keeping the evolution process auditable.

The admitted revisions are then applied to the current manual to form a provisional manual:
\begin{equation}
\widetilde{M}_w \;=\; \mathcal{U}\Big(M_{w-1},
\big\{\, \rho \in \Delta_w
: \mathcal{G}(r_\rho) = 1 \,\big\}\Big),
\label{eq:update}
\end{equation}
where $\mathcal{U}$ applies the admitted revisions and $r_\rho$ denotes the rule introduced or modified by revision $\rho$. 
Although each incorporated revision is individually admissible, multiple revisions may still interact negatively after accumulation. 
The provisional manual is therefore further evaluated by the stability controls described next before becoming the delivered manual $M_w$.

\subsection{Stability-Constrained Skill Evolution}
The admission gate ensures that individual candidate rules are visually grounded and executable, but admissible revisions do not necessarily yield a better manual after accumulation. 
Multiple revisions may interact unexpectedly or gradually introduce noise into the skill library. 
We therefore evaluate each provisional manual on a held-out probe set $P$ before delivering an update. 
The probe set is stratified by reasoning category and remains disjoint from the evolution stream. 
It is never used for reflection or revision. 
The probe score of a manual is defined as:
\begin{equation}
A(M) \;=\; \frac{100}{|P|}
\sum_{x \in P}
\mathbf{1}\big\{\, \hat{y}_{M}(x) = y(x) \,\big\},
\label{eq:probe}
\end{equation}
where $\hat{y}_{M}(x)$ denotes the executor prediction under manual $M$, $y(x)$ is the answer option, and $\mathbf{1}\{\cdot\}$ is the indicator function.

Manual-level evaluation alone, however, does not prevent instability during repeated evolution. 
As rules accumulate, the manual may reach its capacity and require compression. 
Blind compression can remove useful rules together with noisy ones, creating new failures and further revisions. 
We characterize this pathological grow-and-compress regime as:
\begin{equation}
\lim_{W \to \infty} \frac{1}{W}
\sum_{w=1}^{W} \mathbf{1}\{R_w\} \;=\; 1
\qquad\text{and}\qquad
\mathbb{E}\big[A(M_W)\big] \;<\; A(M_0),
\label{eq:mrd}
\end{equation}
where $R_w$ denotes the event that compression is triggered in window $w$. 
To control this failure pattern, SpatialSkill uses three complementary controls: a Growth Budget, Seniority Protection, and a Probe-Score Ratchet. 
They respectively limit manual expansion, protect long-standing rules during compression, and determine whether the resulting update is delivered.

First, excessive growth increases the need for later compression. 
We impose a net-growth budget:
\begin{equation}
|M_w| \;-\; |M_{w-1}| \;\leq\; \beta,
\label{eq:budget}
\end{equation}
where $|M|$ is the number of rules and $\beta$ is the maximum net number of rules added in one window. 
This constraint limits excessive rule accumulation before compression.

Second, compression should not discard long-standing rules indiscriminately. 
We therefore protect stable rules from blind removal. 
A rule that survives multiple windows remains in the next manual unless an evidence-targeted revision explicitly modifies it:
\begin{equation}
r \in M_{w-1},\ \
\mathrm{age}(r) \geq \tau,\ \
r \notin \mathcal{T}_w
\quad\Longrightarrow\quad
r \in M_w,
\label{eq:protection}
\end{equation}
where $\mathrm{age}(r)$ is the number of consecutive windows survived by rule $r$, $\tau$ is the protection window, and $\mathcal{T}_w$ is the set of rules explicitly targeted by evidence-based revisions in window $w$. 
Thus, long-standing rules are protected from blind compression, while evidence-targeted revisions remain possible.

Finally, controlling growth and compression does not guarantee that the resulting manual improves the executor as a whole. 
We therefore use a probe-score ratchet to determine whether a provisional update is delivered. 
Let $b_0=A(M_0)$ denote the initial probe score:
\begin{equation}
M_w =
\begin{cases}
\widetilde{M}_w, & A(\widetilde{M}_w) \geq b_{w-1} - \varepsilon,\\[2pt]
M_{w-1}, & \text{otherwise},
\end{cases}
\qquad
b_w \;=\; \max\big\{\, b_{w-1},\ A(M_w) \,\big\},
\label{eq:ratchet}
\end{equation}
where $\widetilde{M}_w$ denote the constrained provisional update passed to the ratchet, $b_w$ records the best accepted probe score, and $\varepsilon$ absorbs evaluation noise. 
Updates that exceed this tolerance are rejected, while the historical best score is preserved.
Together, these constraints yield the following tolerance-band invariant:
\begin{equation}
A(M_w) \ \geq\ A^{\star}_w - \varepsilon,
\qquad
A^{\star}_w \;=\; \max_{t \le w}\, A(M_t).
\label{eq:invariant}
\end{equation}
where $A^{\star}_w$ denotes the best probe score achieved by any accepted manual up to window $w$. 
Therefore, evolution may plateau or fluctuate within the tolerance range, but large probe-score regressions are not delivered.

\subsection{Category-Conditioned Skill Deployment}
The final skill manual is not uniformly beneficial across reasoning categories. A strategy that helps one category may provide limited benefit for another. We therefore select the inference strategy according to the reasoning category instead of applying the manual globally.

Given the precomputed reasoning category $c \in \mathcal{C}$, the router selects a strategy $s=(i,j)$ from $\mathcal{S}=\{0,1\}^{2}$, where $i=0$ denotes direct answering, $i=1$ denotes chain-of-thought reasoning, $j=1$ denotes providing the final manual $M_W$, and $j=0$ denotes inference without the manual. The routing policy is a mapping $\pi:\mathcal{C}\rightarrow\mathcal{S}$.
If the performance of each strategy were known exactly, the optimal routing policy would maximize expected accuracy:
\begin{equation}
\pi^{\star} \;=\; \arg\max_{\pi} \
\sum_{c \in \mathcal{C}} p(c)\,
\mathrm{Acc}\big(c,\pi(c)\big),
\label{eq:oracle}
\end{equation}
where $p(c)$ is the category prior and $\mathrm{Acc}(c,s)$ denotes the accuracy of strategy $s$ on category $c$. In practice, these quantities are estimated from a finite probe set. To reduce sensitivity to sampling noise, we switch strategies when improvement satisfies a conservative confidence-bound criterion.

Let $\hat{a}_{c,s}$ denote the empirical accuracy of strategy $s$ on category $c$. Using a Wilson confidence interval $[\ell_s(c),u_s(c)]$, the router selects a candidate strategy only when its lower bound exceeds the default strategy's upper bound by a margin $\delta$:
\begin{equation}
\pi(c) \;=\;
\arg\max_{s \in \mathcal{F}(c)}
\hat{a}_{c,s},
\qquad
\mathcal{F}(c)=
\Big\{s\in\mathcal{S}:
\ell_s(c)-u_{s_0(c)}(c)\geq\delta
\Big\},
\label{eq:rule}
\end{equation}
where $s_0(c)$ is the stronger of the two non-manual strategies on the probe set.
If $\mathcal{F}(c)$ is empty, the router retains the default strategy. 
This conservative rule avoids unnecessary switching caused by probe variance.

\section{Experiments}
\subsection{Experiment Setup}
We evaluate on CityCube~\citep{citycube} with its official 4{,}494/528 train/validation split. 
Validation is used only for final reporting. 
The 200-example training-split probe is disjoint from evolution and supplies acceptance and routing decisions. 
At test time, routing receives the pre-annotated CvSI category, we report category-wise and overall micro accuracy.
We compare each frozen executor with its matched no-skill version: Qwen3.5-4B, Qwen3.5-9B, Gemma-3-4B, and GPT-5.6-luna.

\subsection{Implement Details}
The treated executors are Qwen3.5-4B, Qwen3.5-9B, Gemma-3-4B, and GPT-5.6-luna. 
The open-source reference models in the main table are Qwen3-VL-4B/8B/32B \citep{qwen3vl}, Qwen3.5-2B/27B/35B-A3B \citep{qwen3.5}, InternVL3-8B/14B \citep{intervl3}, Gemma-4-31B/26B-A4B, Gemma-3-12B \citep{gemma4}, and Skywork-VL-Reward-7B \citep{DBLP:journals/corr/abs-2505-07263}. 
We also evaluate the open-source spatial models Spatial-SSRL-7B \citep{Spatial-SSRL}, SpaceR-SFT-7B \citep{spacer}, Spatial-MLLM-subset-sft \citep{spatialmllm}, and ViLaSR \citep{vilasr}.
The cross-model transfer analysis additionally uses Qwen3.5-2B as a target executor. 
The proprietary reference baselines are GPT-5.4, GPT-5.6-sol, Gemini-3.5-Flash, Gemini-3.7-Flash, Claude-Opus-5, and Claude-Sonnet-5; GPT-5.6-luna is also evaluated as a treated executor.

The executor and reflection model are frozen throughout. 
The reflection model is Qwen3.5-9B and is served locally for candidate generation, admission checks, and manual revision. 
Open-source executors are served with vLLM on a machine with four NVIDIA RTX 5090 GPUs. 
Smaller models use independent single-GPU replicas; 12B--14B models use tensor parallelism of two; and models of 26B or larger use a four-GPU tensor-parallel server. 
GPU memory utilization is set to 0.95, each request accepts up to ten images, and each server processes one concurrent multimodal sequence. 
Local runs use greedy decoding, while API baselines use their fixed native serving configuration and the same answer-extraction rule.

\subsection{Main Results}

\begin{table}[t]
  \centering
  \scriptsize
  \caption{Main results on the CityCube dataset.
  Accuracy is reported in percent.
  Option-wise Acc. Range (pp) is the maximum difference in conditional accuracy across predicted answer options.}
  \label{maintab}
  \begingroup
  \setlength{\tabcolsep}{3.6pt}
  \renewcommand{\arraystretch}{1.1}
  \begin{tabular}{lrrrrrrrr}
    \toprule
    Method
      & \multicolumn{7}{c}{Accuracy (\%) $\uparrow$}
      & Option-wise Acc. Range (pp) $\downarrow$ \\
    \cline{2-8}
      & Overall & $\Delta$ & CR & MR & PT & SR & WK & \\
    \midrule
    Human
      & 88.3 & -- & 93.1 & 92.4 & 87.4 & 90.2 & 78.6 & -- \\ \midrule
    \multicolumn{9}{l}{\textit{Paired executors}} \\
    \midrule
    \textbf{Qwen3.5-4B + SpatialSkill}
      & \textbf{52.08} & $+4.92$ & 43.24 & 51.22 & 41.77 & 56.10 & 64.06 & 23.52 \\
      \quad Qwen3.5-4B
      & 47.16 & -- & 51.35 & 53.66 & 39.87 & 45.53 & 52.34 & 27.28 \\
    \hdashline
    \textbf{Qwen3.5-9B + SpatialSkill}
      & \textbf{57.58} & $+3.03$ & 59.46 & 67.07 & 51.90 & 52.85 & 62.50 & 23.10 \\
      \quad Qwen3.5-9B
      & 54.55 & -- & 56.76 & 64.63 & 47.47 & 47.97 & 62.50 & 39.21 \\
    \hdashline
    \textbf{Gemma-3-4B + SpatialSkill}
      & \textbf{51.89} & $+2.65$ & 51.35 & 60.98 & 41.77 & 52.03 & 58.59 & 47.94 \\
      \quad Gemma-3-4B
      & 49.24 & -- & 51.35 & 59.76 & 37.97 & 52.03 & 53.12 & 72.12 \\
    \hdashline
    \textbf{GPT-5.6-luna + SpatialSkill}
      & \textbf{56.82} & $+3.03$ & 62.16 & 62.20 & 45.57 & 56.91 & 65.62 & 19.37 \\
      \quad GPT-5.6-luna
      & 53.79 & -- & 48.65 & 57.32 & 42.41 & 59.35 & 61.72 & 30.43 \\
    \midrule
    \multicolumn{9}{l}{\textit{Open-source reference models}} \\
    \midrule
    Qwen3-VL-4B
      & 53.03 & -- & 56.76 & 59.76 & 41.14 & 56.91 & 58.59 & 24.93 \\
    Qwen3-VL-8B
      & 50.95 & -- & 45.95 & 53.66 & 39.24 & 51.22 & 64.84 & 24.96 \\
    Qwen3-VL-32B
      & 53.79 & -- & 51.35 & 59.76 & 41.77 & 54.47 & 64.84 & 18.04 \\
    Qwen3.5-2B 
      & 46.59 & -- & 48.65 & 53.66 & 33.54 & 52.85 & 51.56 & 49.77 \\
    Qwen3.5-27B
      & 58.14 & -- & 51.35 & 64.63 & 45.57 & 56.10 & 73.44 & 17.36 \\
    Qwen3.5-35B-A3B
      & 53.22 & -- & 56.76 & 54.88 & 41.77 & 55.28 & 63.28 & 31.00 \\
    InternVL3-8B
      & 49.43 & -- & 48.65 & 47.56 & 43.67 & 49.59 & 57.81 & 20.14 \\
    InternVL3-14B
      & 51.70 & -- & 56.76 & 51.22 & 44.30 & 52.03 & 59.38 & 51.97 \\
    Gemma-4-31B
      & 54.55 & -- & 40.54 & 53.66 & 41.77 & 56.10 & 73.44 & 26.52 \\
    Gemma-4-26B-A4B
      & 52.27 & -- & 40.54 & 56.10 & 41.14 & 56.91 & 62.50 & 17.78 \\
    Gemma-3-12B
      & 50.95 & -- & 56.76 & 54.88 & 39.87 & 47.15 & 64.06 & 39.70 \\
    Skywork-VL-Reward-7B
      & 45.83 & -- & 37.84 & 45.12 & 34.81 & 51.22 & 57.03 & 28.53 \\
    \midrule
    \multicolumn{9}{l}{\textit{Open-source Spatial models}} \\
    \midrule
    Spatial-SSRL-7B & 46.97 & -- & 45.95 & 46.34 & 34.81 & 50.41 & 59.38 & 13.76 \\
    SpaceR-SFT-7B & 44.13 & -- & 54.05 & 47.56 & 36.71 & 33.33 & 58.59 & 41.07 \\
    Spatial-MLLM-subset-sft& 39.77 & -- & 37.84 & 36.59 & 36.71 & 39.02 & 46.88 & 73.94 \\
    ViLaSR& 49.81 & -- & 51.35 & 56.1 & 36.08 & 47.97 & 64.06 & 26.79 \\
    \midrule
    \multicolumn{9}{l}{\textit{Closed-source reference models}} \\
    \midrule
    GPT-5.4
      & 50.38 & -- & 37.84 & 57.32 & 41.14 & 50.41 & 60.94 & 31.00 \\
    Gemini-3.5-Flash
      & 48.11 & -- & 40.54 & 53.66 & 39.87 & 50.41 & 54.69 & 16.95 \\
    GPT-5.6-sol
      & 56.06 & -- & 45.95 & 67.07 & 44.30 & 54.47 & 67.97 & 24.36 \\
    Qwen3.7-plus
      & 56.06 & -- & 48.65 & 62.20 & 46.20 & 56.91 & 65.62 & 12.96 \\
    Gemini-3.7-Flash
      & 52.27 & -- & 45.95 & 57.32 & 43.67 & 53.66 & 60.16 & 14.28 \\
    Claude-Sonnet-5
      & 52.65 & -- & 45.95 & 50.00 & 49.37 & 48.78 & 64.06 & 24.05 \\
    Claude-Opus-5
      & 52.08 & -- & 35.14 & 60.98 & 44.30 & 48.78 & 64.06 & 20.67 \\
    \bottomrule
  \end{tabular}%
  \endgroup
  \vspace{-6mm}
\end{table}

Table~\ref{maintab} reports the main results on CityCube \citep{citycube}.
We focus on paired comparisons between each frozen executor and its SpatialSkill-enhanced counterpart, while also reporting stronger open-source, spatially specialized, and closed-source models as reference points.

\textbf{SpatialSkill consistently improves frozen executors.}
Across all four paired executors, SpatialSkill improves overall accuracy by 2.65--4.92 percentage points, with a mean gain of 3.41 points.
Qwen3.5-4B improves from 47.16\% to 52.08\%, Qwen3.5-9B from 54.55\% to 57.58\%, Gemma-3-4B from 49.24\% to 51.89\%, and GPT-5.6-luna from 53.79\% to 56.82\%.
Because the executor weights remain frozen throughout, these gains are achieved through explicit skill evolution and prompt-level deployment rather than parameter updates.
The resulting knowledge remains stored in a natural-language manual, making the acquired reasoning strategies explicit, inspectable, and retractable.

\textbf{SpatialSkill narrows the gap to stronger reference models.}
Qwen3.5-9B equipped with SpatialSkill reaches 57.58\%, only 0.56 percentage points below the larger Qwen3.5-27B baseline at 58.14\%.
It also exceeds the best-performing closed-source reference models in our evaluation, GPT-5.6-sol and Qwen3.7-plus, both at 56.06\%, by 1.52 points.
Although Qwen3.5-27B remains the best-performing non-human model overall, these results show that explicit skill evolution can recover part of the performance gap to stronger models without modifying the executor backbone.

\textbf{The gains are category- and executor-dependent.}
For Qwen3.5-4B, SpatialSkill improves SR and WK by 10.57 and 11.72 percentage points, respectively, but decreases performance on CR and MR.
GPT-5.6-luna gains 13.51 points on CR but slightly declines on SR, while Qwen3.5-9B improves or matches its baseline across all five categories and Gemma-3-4B benefits mainly on MR, PT, and WK.
This heterogeneity shows that skill effectiveness is not uniform across reasoning categories or executors, supporting category-conditioned routing rather than a single global deployment strategy.

\textbf{A substantial gap remains between current models and human performance.}
For context, CityCube reports an overall human accuracy of 88.3\%, whereas the best-performing non-human model in our evaluation, Qwen3.5-27B, reaches 58.14\%, leaving a gap of 30.2 percentage points.
The gap is particularly pronounced for perspective taking (PT), where the best non-human result is 51.9\% compared with 87.4\% for humans, but is much smaller for world knowledge (WK), at 73.44\% versus 78.6\%.
This pattern suggests that current models remain particularly weak on perspective-dependent operations involving viewpoint transformation and cross-view alignment.
Moreover, all four evaluated open-source spatial models remain below 50\% overall accuracy, indicating that existing spatially specialized models still leave substantial room for improvement on robust cross-view reasoning.

\textbf{The gains are not accompanied by larger option-dependent disparity.}
As a descriptive diagnostic, we report the option-wise accuracy range, defined as the maximum difference in conditional accuracy across predicted answer options.
Across all four paired executors, SpatialSkill reduces this range: from 27.28 to 23.52 for Qwen3.5-4B, 39.21 to 23.10 for Qwen3.5-9B, 72.12 to 47.94 for Gemma-3-4B, and 30.43 to 19.37 for GPT-5.6-luna.
Thus, the overall accuracy gains are not accompanied by greater option-dependent disparity.

\begin{wrapfigure}{r}{0.45\textwidth}
  \centering
  \footnotesize
  \captionof{table}{Overall ablations validation accuracy (\%) on Qwen3.5-4B and Gemma-3-4B.}
  \setlength{\tabcolsep}{3pt}
  \begin{tabular}{lrr}
    \toprule
    Condition & Qwen & Gemma \\
    \midrule
    Direct baseline & 47.16 & 49.24 \\
    CoT baseline & 49.62 & 48.67 \\
    \midrule
    w/o attribution-aware reflection & 50.00 & 47.35 \\
    w/o visual-evidence criterion
      & 50.76
      & 50.74 \\
    w/o executability criterion
      & 51.14
      & 51.32 \\
    w/o joint admission gate & 50.19 & 50.95 \\
    w/o growth budget
      & 51.57
      & 51.31 \\
    w/o seniority protection
      & 51.68
      & 51.44 \\
    w/o probe-score ratchet & 51.14 & 50.57 \\
    \midrule
    \textsc{SpatialSkill} & 52.08 & 51.89 \\
    \bottomrule
  \end{tabular}
  \label{tab:component-ablation}
\end{wrapfigure}

\subsection{Ablation Experiments}
Table~\ref{tab:component-ablation} reports results under the same frozen executor.
All ablation settings use the same frozen executor, decoding parameters, validation split, and answer-extraction procedure as the complete system. 
Evolution-based variants restart from the empty manual and produce a terminal artifact that is fixed before validation.

The Direct baseline answers without a manual or chain-of-thought prompting. 
The CoT baseline uses chain-of-thought prompting without a manual. 
SpatialSkill uses the final manual with the category-conditioned policy.
The component variants remove one named mechanism at a time:
  \emph{w/o attribution-aware reflection} removes the separation between reusable skill-level feedback and execution-level notes;
  \emph{w/o visual-evidence criterion} removes $\mathcal{V}$ while retaining $\mathcal{E}$;
  \emph{w/o executability criterion} removes $\mathcal{E}$ while retaining $\mathcal{V}$;
  \emph{w/o joint admission gate} removes both predicates from $\mathcal{G}(r)$;
  \emph{w/o growth budget} removes the per-window net-growth constraint $\beta$;
  \emph{w/o seniority protection} removes the protection threshold $\tau$; and
  \emph{w/o probe-score ratchet} removes the probe-based acceptance rule and is referred to as \emph{Naive} in the stability analysis.


\textbf{Learned skills improve over no-skill baselines.}
The direct baseline reaches 47.16\% on Qwen3.5-4B and 49.24\% on Gemma-3-4B, while the CoT baseline reaches 49.62\% and 48.67\% without a manual. 
SpatialSkill reaches 52.08\% and 51.89\%, yielding gains of 4.92 and 2.65 percentage points over the corresponding Direct baselines. 
The gap between CoT and SpatialSkill shows that the improvement cannot be explained by CoT prompting alone.

\textbf{Attribution-aware reflection is important for reliable skill induction.}
Removing attribution-aware reflection reduces accuracy to 50.00\% on Qwen3.5-4B and 47.35\% on Gemma-3-4B, a loss of 2.08 and 4.54 points from the complete system. 
The larger degradation on Gemma suggests that separating reusable and execution-specific feedback is particularly important for this executor.
The result suggests that separating reusable skills from execution-specific notes is important for stable reflection.

\textbf{Grounded admission benefits from both evidence and executability checks.}
Removing the visual-evidence criterion gives 50.76\% and 50.74\%, whereas removing the executability criterion gives 51.14\% and 51.32\% on Qwen3.5-4B and Gemma-3-4B, respectively.
Removing the joint admission gate gives 50.19\% and 50.95\%.
All three variants underperform the complete system on both executors.
On Qwen3.5-4B, removing both criteria is more harmful than removing either criterion alone; on Gemma-3-4B, the effects are not strictly additive.
These results support the utility of grounded admission while also showing that the relative contribution of the two criteria is executor-dependent.

\textbf{Stability controls provide complementary protection.}
Removing growth budget reduces accuracy to 51.57\% and 51.31\%, while removing seniority protection gives 51.68\% and 51.44\%. 
Removing probe-score ratchet produces larger drops of 0.94 points on Qwen3.5-4B and 1.32 points on Gemma-3-4B, reaching 51.14\% and 50.57\%. 
This pattern suggests that the ratchet serves as the primary safeguard against harmful revisions, whereas budget and seniority constraints regulate the trajectory prior to acceptance. 
Their effects are therefore complementary rather than interchangeable.

\begin{wrapfigure}{r}{0.58\textwidth}
  \centering
  \footnotesize
  \captionof{table}{Routing summary and paired error turnover on the held-out validation split.}
  \label{tab:routing-error}
  \setlength{\tabcolsep}{3pt}
  \begin{tabular}{lrrrrrr}
    \toprule
    \multicolumn{7}{c}{\textbf{(a) Routing summary}} \\
    \midrule
    \multicolumn{7}{c}{%
      \begin{tabular}{@{}lrrrr@{}}
        Executor & Main & Probe replay & No-margin & Oracle \\
        \midrule
        Qwen3.5-4B  & 52.08 & 53.03 & 51.33 & 54.17 \\
        Gemma-3-4B & 51.89 & 50.95 & 51.33 & 53.03 \\
      \end{tabular}
    } \\
    \midrule
    \multicolumn{7}{c}{\textbf{(b) Paired error turnover}} \\
    \midrule
    Executor & Base & Main & Fixed & Broken & Net & Persistent \\
    \midrule
    Qwen3.5-4B  & 47.16 & 52.08 & 87 & 61 & $+26$ & 192 \\
    Gemma-3-4B & 49.24 & 51.89 & 70 & 56 & $+14$ & 198 \\
    \bottomrule
  \end{tabular}
\end{wrapfigure}

\subsection{Stability of Skill Evolution}
Perceptual spatial reasoning has no symbolic execution oracle, so we measure stability over 43-window trajectories from the same frozen executor and training stream.
A refactor event counts when the manual is explicitly refactored or shrinks between adjacent windows.
Naive denotes the ablation variant following the same skill-evolution loop but removing the probe-score ratchet, lacking the safeguard against harmful revision deployment.

\begin{figure}[t]
  \centering
  \begin{subfigure}[t]{0.32\columnwidth}
    \centering
    \includegraphics[width=\linewidth]{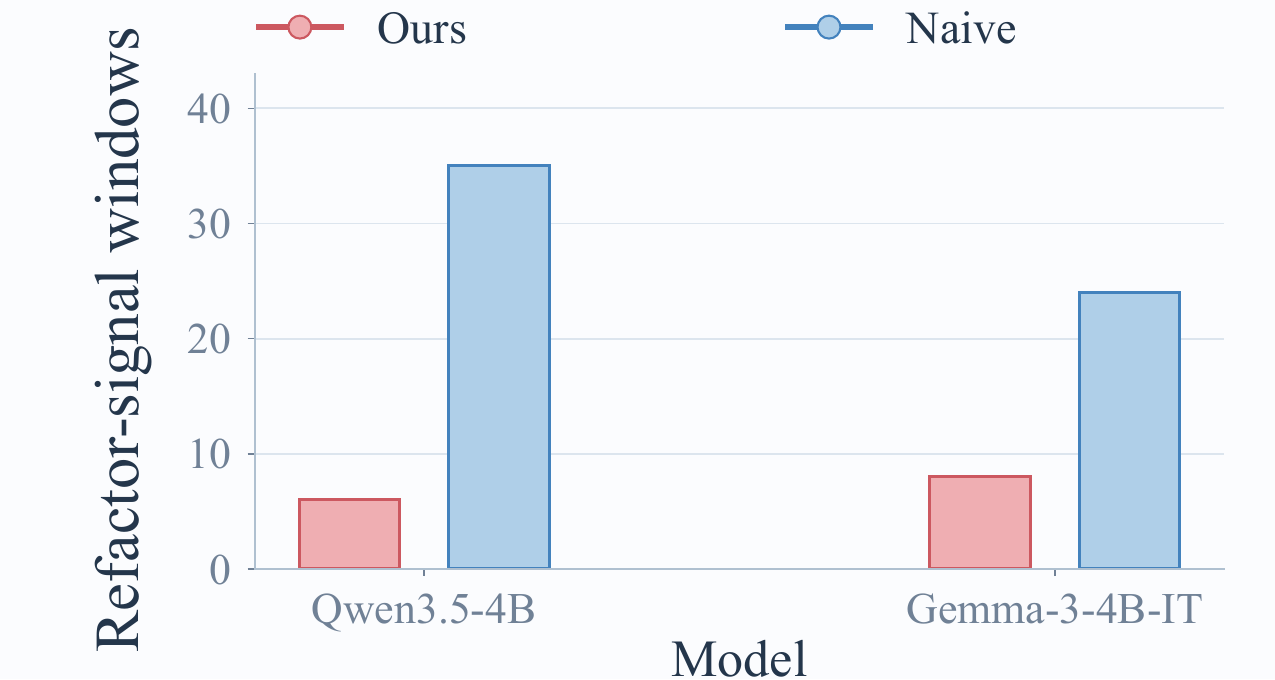}
    \caption{Refactor-event windows.}
    \label{fig:e3}
  \end{subfigure}\hfill
  \begin{subfigure}[t]{0.32\columnwidth}
    \centering
    \includegraphics[width=\linewidth]{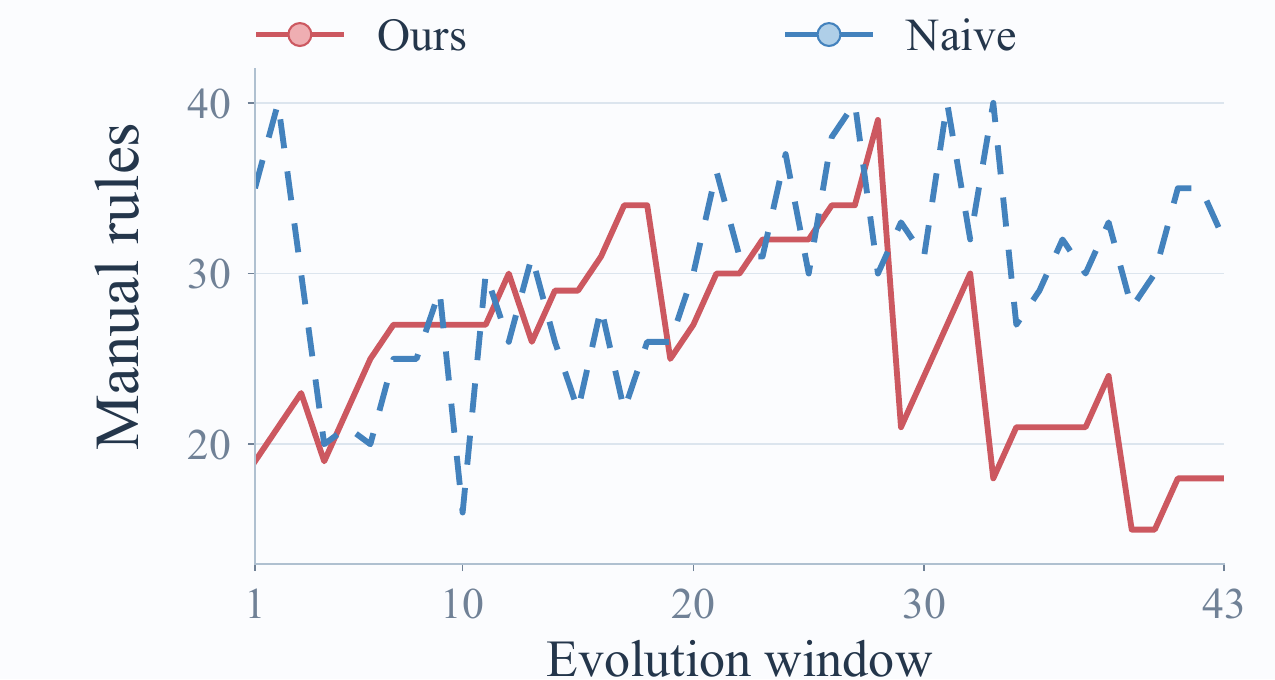}
    \caption{Qwen3.5-4B manual size.}
    \label{fig:e3_qwen}
  \end{subfigure}\hfill
  \begin{subfigure}[t]{0.32\columnwidth}
    \centering
    \includegraphics[width=\linewidth]{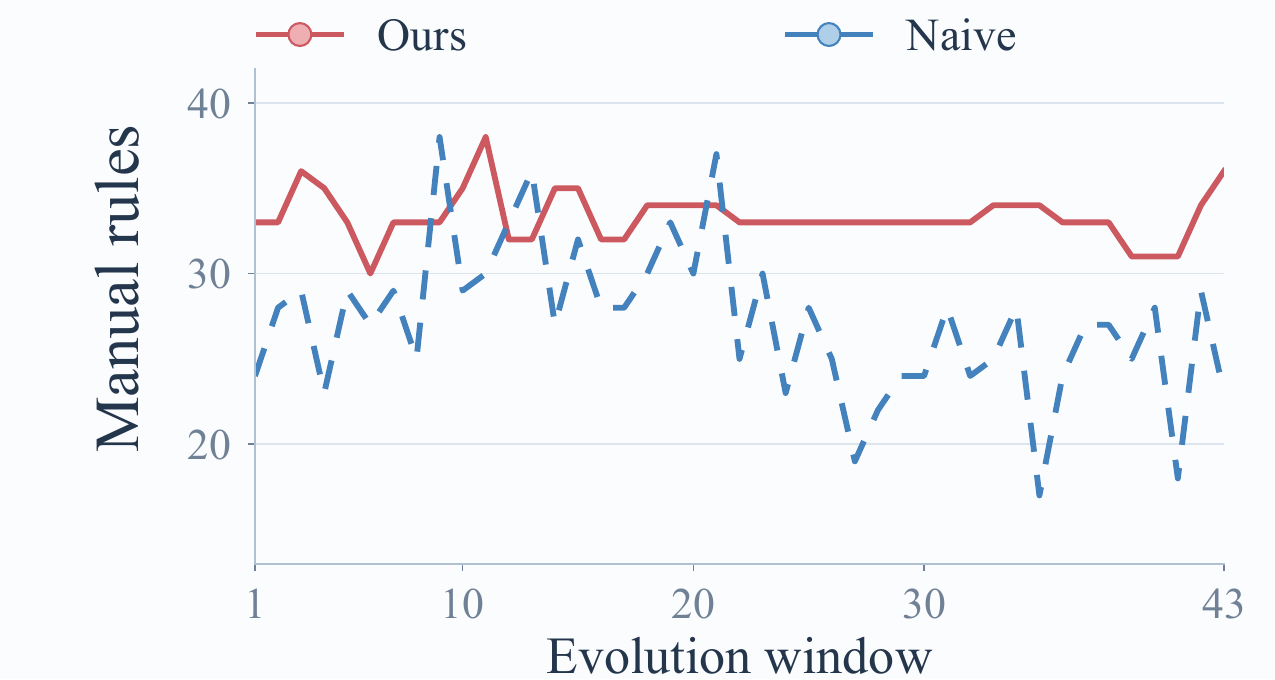}
    \caption{Gemma-3-4B manual size.}
    \label{fig:e3_gemma}
  \end{subfigure}
  \caption{Stability diagnostics over 43 evolution windows for SpatialSkill and Naive.}
  \label{fig:e3_stability}
\end{figure}

\textbf{The ratchet blocks harmful revisions and reduces unstable refactors.}
SpatialSkill reduces refactor-event windows from 35 to 6 on Qwen3.5-4B and from 24 to 8 on Gemma-3-4B in Fig.~\ref{fig:e3}.
The ratchet rejects 14 and 21 revisions, all of which score lower than the accepted reference, with worst drops of 9.0 and 5.5 points.
In the evaluated trajectories, no accepted revision decreases the probe score.
Manual-size contractions occur in 17 and 16 Naive windows, compared with 6 and 8 SpatialSkill windows shown in Figs.~\ref{fig:e3_qwen}--\ref{fig:e3_gemma}.

\textbf{Stability translates into better held-out validation.}
SpatialSkill reaches 52.08\% versus 51.14\% for Naive on Qwen3.5-4B, and 51.89\% versus 50.57\% on Gemma-3-4B.
Thus, the safeguards do not merely suppress change. 
The joint admission gate screens candidate rules, while the retained growth and protection controls constrain the trajectory.
The ratchet prevents deploying revisions whose probe score falls beyond the tolerance band.
Safe compression remains allowed, but harmful revisions are not deployed.

\subsection{Skill-Aware Routing Analysis}
\begin{figure}[t]
  \centering
  \begin{subfigure}[t]{0.48\columnwidth}
    \centering
    \includegraphics[width=\linewidth]{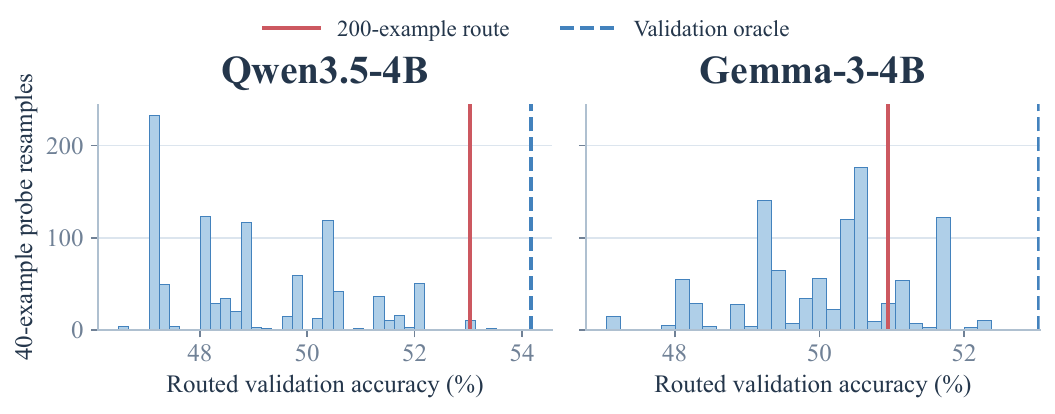}
    \caption{Sensitivity under resampled probes.}
    \label{fig:e4_validation}
  \end{subfigure}\hfill
  \begin{subfigure}[t]{0.48\columnwidth}
    \centering
    \includegraphics[width=\linewidth]{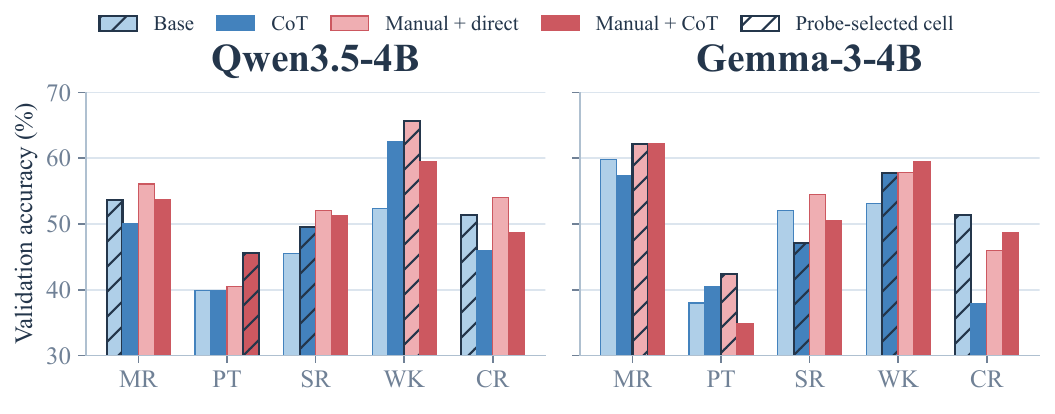}
    \caption{Validation cells with probe selections hatched.}
    \label{fig:e4_small_probe}
  \end{subfigure}
  \caption{Probe-derived routing diagnostics: full-probe route and validation oracle.}
  \label{fig:e4_routing}
\end{figure}

\textbf{Routing balances empirical performance and conservative deployment.}
The analysis-only probe replay reaches 53.03\% for Qwen3.5-4B and 50.95\% for Gemma-3-4B, compared with the Main results of 52.08\% and 51.89\%.
The No-margin diagnostic uses the same probe-based routing rule with the switching margin set to $\delta=0$, reaching 51.33\% on both executors.
The non-deployable validation oracle reaches 54.17\% and 53.03\% in Table~\ref{tab:routing-error} (a).
Probe replay applies a fixed finite-probe route to the bank of validation predictions.
It diagnoses sensitivity to route estimation.
The lower No-margin accuracy suggests that conservative switching can reduce over-selection under finite-probe uncertainty.
Moreover, routed accuracy varies widely across 1,000 resampled 40-example probes in Fig.~\ref{fig:e4_validation}, which further motivates the default-and-margin rule: under insufficient evidence, the router retains the default instead of selecting a noise-inflated winner.


\textbf{Optimal delivery is family- and executor-dependent.}
Qwen3.5-4B favors manual+direct on MR, SR, and WK, and manual+CoT on PT. Gemma-3-4B favors manual+direct on PT and SR, manual+CoT on WK, and the base configuration on CR, as shown in Fig.~\ref{fig:e4_small_probe}.
The hatched probe-selected cells need not equal the validation-best cells, because validation labels never enter policy construction.
This variation supports the multi-referential heterogeneity premise: delivery utility depends on cognitive family and executor.

\subsection{Paired Error Analysis}
\begin{figure}[t]
  \centering
  \begin{subfigure}[t]{0.4\linewidth}
    \centering
    \includegraphics[width=\linewidth]{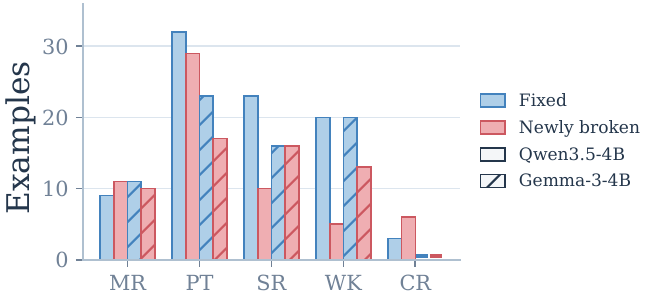}
    \caption{Error turnover by CvSI category.}
    \label{fig:e5-1}
  \end{subfigure}
  \hspace{0.02\linewidth}
  \begin{subfigure}[t]{0.4\linewidth}
    \centering
    \includegraphics[width=\linewidth]{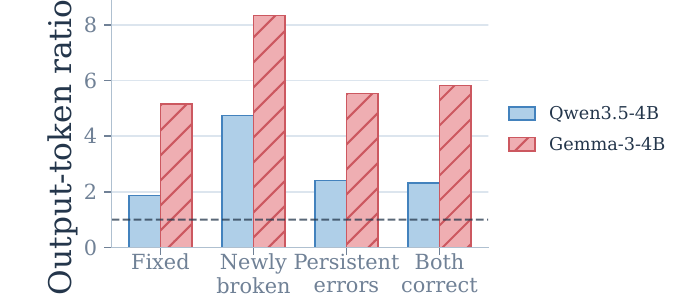}
    \caption{Output-token expansion by transition.}
    \label{fig:e5-2}
  \end{subfigure}
  \caption{Paired error-turnover diagnostics for the frozen predictions.}
  \label{fig:e5-error-turnover}
\end{figure}

\textbf{Gains come from targeted repairs rather than uniform correction.}
We froze base and SpatialSkill predictions on the same validation examples. 
Qwen3.5-4B repairs 87 baseline errors and introduces 61 new errors, for a net gain of 26. 
Gemma-3-4B repairs 70 and breaks 56, for a net gain of 14 shown in Table~\ref{tab:routing-error} (b).
Repairs concentrate in WK and SR for Qwen and in WK and PT for Gemma, whereas 192 and 198 baseline errors persist in Fig.~\ref{fig:e5-1}.
Thus, the evolved manual improves aggregate accuracy through targeted repairs rather than uniform correction of the baseline error set.

\textbf{Longer outputs do not indicate correct answers.}
Conditioned on transition type, the mean SpatialSkill-to-base output-token ratio is largest for newly broken examples (4.74$\times$ for Qwen and 8.33$\times$ for Gemma), and remains elevated for fixed examples, as shown in Fig.~\ref{fig:e5-2}.
Length expansion is therefore only a descriptive correlate of the applied reasoning process rather than a reliable indicator of correctness.

\subsection{Cross-Model Transfer with Source-Routed Delivery}
\label{app:transfer}

We evaluate whether a manual and its category-level delivery policy can transfer to a different executor without target-specific adaptation. 
For each pair, the target executor, validation split, decoding configuration, and metric remain fixed; the source manual and source-derived routing policy are applied directly to the target. 
Each pair uses a pre-specified terminal source artifact, and no transfer round is selected using held-out validation performance.

\begin{figure}[h]
  \centering
  \includegraphics[width=0.72\textwidth]{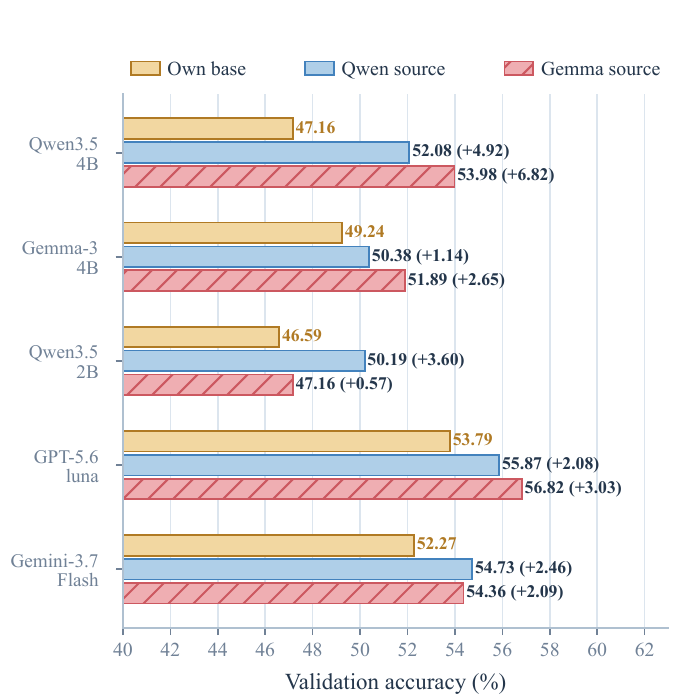}
  \caption{Own-base and cross-model source-routed validation accuracy. Each transferred manual and its source-derived routing policy is applied to the target executor without target-specific adaptation; N/A denotes an incomplete pairing.}
  \label{fig:e6}
\end{figure}

All eight completed cross-model transfers improve over their target bases, with gains ranging from 0.57 to 6.82 percentage points and a mean gain of 2.72 points. 
Gemma-source gives the largest gain on Qwen3.5-4B, improving from 47.16\% to 53.98\%, while Qwen-source is strongest on Qwen3.5-2B, improving from 46.59\% to 50.19\%. 
GPT-5.6-luna reaches 55.87\% with Qwen-source and 56.82\% with Gemma-source. 
Gemini-3.7-Flash reaches 54.73\% and 54.36\% with the two source manuals, respectively. 
The results indicate that source-routed skills can transfer across executor families, but neither source manual is uniformly optimal.

\subsection{Skill Manual Transfer Experiment}
Our external transfer set originates from MMSI-Bench \citep{mmsibench} as the second dataset.
After data cleaning, we use the 1,000 instances for transfer evaluation.
We evaluate the frozen executor-specific manual and routing policy without adapting the model or manual.

\begin{table}[h]
  \centering
  \footnotesize
  \caption{Own-base and frozen manual-routed MMSI-Bench accuracy and option-wise accuracy range.
  Each executor's frozen manual and source-derived routing policy is applied to the 1,000-example transfer set .}
  \label{tab-transfer}
  \begingroup
  \setlength{\tabcolsep}{3pt}
  \renewcommand{\arraystretch}{1.18}
  \begin{tabular}{lrr}
    \hline
    Method & Overall Acc. (\%) $\uparrow$ & Option-wise Acc. Range (pp) $\downarrow$ \\
    \hline
    Qwen3.5-4B Base
      & 29.30 & 10.45 \\
    Qwen3.5-4B Manual + route
      & \textbf{31.30} & \textbf{8.01} \\
    \hdashline
    GPT-5.6-luna Base
      & 34.30 & 10.62 \\
    GPT-5.6-luna Manual + route
      & \textbf{38.40} & \textbf{3.95} \\
    \hline
  \end{tabular}
  \endgroup
\end{table}

This experiment tests transfer across datasets and scene domains rather than transfer between executors.
We retain overall accuracy and the option-wise accuracy range used in Table~\ref{tab-transfer}.

Both frozen manual policy pairs improve over their bases.
Qwen3.5-4B improves by 2.00 points, from 29.30\% to 31.30\%, while its option-wise accuracy range decreases from 10.45 to 8.01 points.
GPT-5.6-luna improves by 4.10 points, from 34.30\% to 38.40\%, while its option-wise accuracy range decreases from 10.62 to 3.95 points.
These single-run external results support the transfer of the frozen delivery artifacts. They further suggest that the learned manual has some transferability to similar spatial-reasoning tasks.

\subsection{Paired Error Analysis}
For Qwen3.5-4B and Gemma-3-4B, base and SpatialSkill predictions are paired on the same validation examples. 
We classify each example as repaired, broken, fixed, or persistent according to the transition from the base prediction to the SpatialSkill prediction. 
Qwen3.5-4B repairs 87 baseline errors and introduces 61 new errors, for a net gain of 26. 
Gemma-3-4B repairs 70 baseline errors and introduces 56 new errors, for a net gain of 14. 
The remaining baseline errors persist in the majority of cases, showing that the manual produces targeted repairs rather than uniform correction.

We also compare generated-token ratios across transition types. 
The largest SpatialSkill-to-base expansion occurs on newly broken examples, with ratios of 4.74 for Qwen3.5-4B and 8.33 for Gemma-3-4B. 
Output length is therefore a descriptive correlate of skill-conditioned reasoning, not evidence of correctness or a causal mechanism.

\subsection{Generality Across Reflection Models}
\label{app:reflection_model}
The main experiments use Qwen3.5-9B as the fixed reflection model for candidate skill generation, admission checking, and manual revision. 
To examine whether SpatialSkill depends on this particular reflector, we keep the executor fixed as Qwen3.5-4B and repeat the complete skill-evolution pipeline with two alternative reflection models: Qwen3-4B and Qwen3-8B. 
All other settings, including the evolution stream, probe split, window size, admission criteria, stability controls, routing procedure, and executor decoding configuration, are kept unchanged. 
Thus, the reflection model is the only experimental variable.

Table~\ref{tab:reflector_generalization} reports the results. 
With the default Qwen3.5-9B reflector, SpatialSkill improves the frozen Qwen3.5-4B executor from 47.16\% to 52.08\%. 
Replacing the reflector with Qwen3-4B and Qwen3-8B yields accuracies of 50.57\% and 49.05\%, corresponding to gains of +3.41 and +1.89 percentage points over the same frozen executor baseline. 
The consistent gains across these reflectors suggest that SpatialSkill is not tied to the default reflection model, although performance remains sensitive to reflector choice.

\begin{table}[t]
\centering
\caption{
Reflection-model sensitivity with Qwen3.5-4B as the frozen executor.
Only the reflection model is changed; all other evolution and inference settings are fixed.
}
\label{tab:reflector_generalization}
\small
\begin{tabular}{lccc}
\toprule
Reflection Model & Executor & Accuracy (\%) $\uparrow$ & $\Delta$ (pp) $\uparrow$ \\
\midrule
None & Qwen3.5-4B & 47.16 & -- \\
\midrule
Qwen3-4B & Qwen3.5-4B & 50.57 & +3.41 \\
Qwen3-8B & Qwen3.5-4B & 49.05 & +1.89 \\
Qwen3.5-9B & Qwen3.5-4B & 52.08 & +4.92 \\
\bottomrule
\end{tabular}
\end{table}


\section{Conclusion}
In this paper, we study the self-evolution of spatial skills in frozen vision-language models and propose \textit{SpatialSkill}, a framework that accumulates explicit natural-language skills without parameter updates.
Our analysis highlights three requirements for reliable perceptual skill evolution: grounding candidate skills, stabilizing repeated accumulation, and selectively deploying skills across reasoning categories.
Experiments on CityCube show that explicit skill evolution can improve frozen executors while reducing instability across repeated evolution. 
We further find that the benefit of skill usage varies across spatial-reasoning categories, motivating category-aware skill routing rather than a uniform manual application strategy. 
The limitation is that current routing relies on precomputed category annotations, and natural-language manuals can improve reasoning strategies but do not modify the executor's underlying visual capabilities. 
Future work will explore the use of explicit manuals as fine-tuning priors to combine explicit and implicit knowledge updates.


\bibliography{iclr2027_conference}
\bibliographystyle{iclr2027_conference}


\end{document}